\documentclass[journal]{IEEEtai}

\usepackage[colorlinks,urlcolor=blue,linkcolor=blue,citecolor=blue]{hyperref}

\usepackage{color,array}

\usepackage{graphicx}

\usepackage{subfig}
\usepackage{url}            
\usepackage{booktabs}       
\usepackage{amsfonts}       
\usepackage{nicefrac}       
\usepackage{microtype}      
\usepackage{xcolor}         
\usepackage{multirow}
\usepackage{graphicx}
\usepackage{cite}
\usepackage{amsmath}
\usepackage{amssymb}
\usepackage{stmaryrd}
\usepackage{capt-of}
\usepackage{arydshln, bm}
\usepackage{colortbl}

\usepackage{xspace}
\makeatletter
\DeclareRobustCommand\onedot{\futurelet\@let@token\@onedot}
\def\@onedot{\ifx\@let@token.\else.\null\fi\xspace}
\def\eg{\emph{e.g}\onedot}

\def\etal{\emph{et al}\onedot}

\newcommand{\Btheta}{{\bm \theta}}
\newcommand{\Bphi}{{\bm \phi}}

\begin{document}

\title{Generalizable Synthetic Image Detection via \\ Language-guided Contrastive Learning} 

\author{Haiwei~Wu,~Jiantao~Zhou,~\IEEEmembership{Senior Member,~IEEE},~and~Shile~Zhang
	\IEEEcompsocitemizethanks{
		\IEEEcompsocthanksitem H. Wu is with the School of Computer Science and Engineering, University of Electronic Science and Technology of China, Chengdu 611731, China. Email: haiweiwu@uestc.edu.cn.
		\IEEEcompsocthanksitem J. Zhou is with the State Key Laboratory of Internet of Things for Smart City, and also with the Department of Computer and Information Science, Faculty of Science and Technology, University of Macau, Macau 999078, China. Email: jtzhou@umac.mo. \emph{(Corresponding author: Jiantao Zhou.)}
		\IEEEcompsocthanksitem S. Zhang is with 
		Faculty of Applied Sciences, Macao Polytechnic University, Macau 999078, China. Email: P2413115@mpu.edu.mo.
	}
}

\maketitle

\begin{abstract}
The heightened realism of AI-generated images can be attributed to the rapid development of synthetic models, including generative adversarial networks (GANs) and diffusion models (DMs). The malevolent use of synthetic images, such as the dissemination of fake news or the creation of fake profiles, however, raises significant concerns regarding the authenticity of images. Though many forensic algorithms have been developed for detecting synthetic images, their performance, especially the generalization capability, is still far from being adequate to cope with the increasing number of synthetic models. In this work, we propose a simple yet very effective synthetic image detection method via a language-guided contrastive learning. Specifically, we augment the training images with carefully-designed textual labels, enabling us to use a joint visual-language contrastive supervision for learning a forensic feature space with better generalization. It is shown that our proposed LanguAge-guided SynThEsis Detection (LASTED) model achieves much improved generalizability to unseen image generation models and delivers promising performance that far exceeds state-of-the-art competitors over four datasets. The code is available at \url{https://github.com/HighwayWu/LASTED}.
\end{abstract}

\begin{IEEEImpStatement}
Synthetic image detection research has become crucial in combating the proliferation of digitally manipulated visuals across media platforms. However, current detection methods have shown limitations in out-of-domain testing scenario, with studies indicating that up to 35\% of synthetic images can evade existing detectors. The innovative method presented in this word address these shortcomings by significantly enhancing detection accuracy to over 95\%. This advancement not only fortifies trust in digital media but also broadens the applications of synthetic image detection to areas such as online content moderation, digital forensics, and the protection of intellectual property. Furthermore, this work could provide a vital tool for educators and policymakers in their efforts to combat misinformation and uphold digital authenticity.
\end{IEEEImpStatement}

\begin{IEEEkeywords}
Artificial intelligence in digital humanity, deep learning, digital image forensics, forgery detection
\end{IEEEkeywords}


\section{Introduction}

\begin{figure}[t!]
	\centering
	\subfloat{
		\includegraphics[width = 0.48\textwidth]{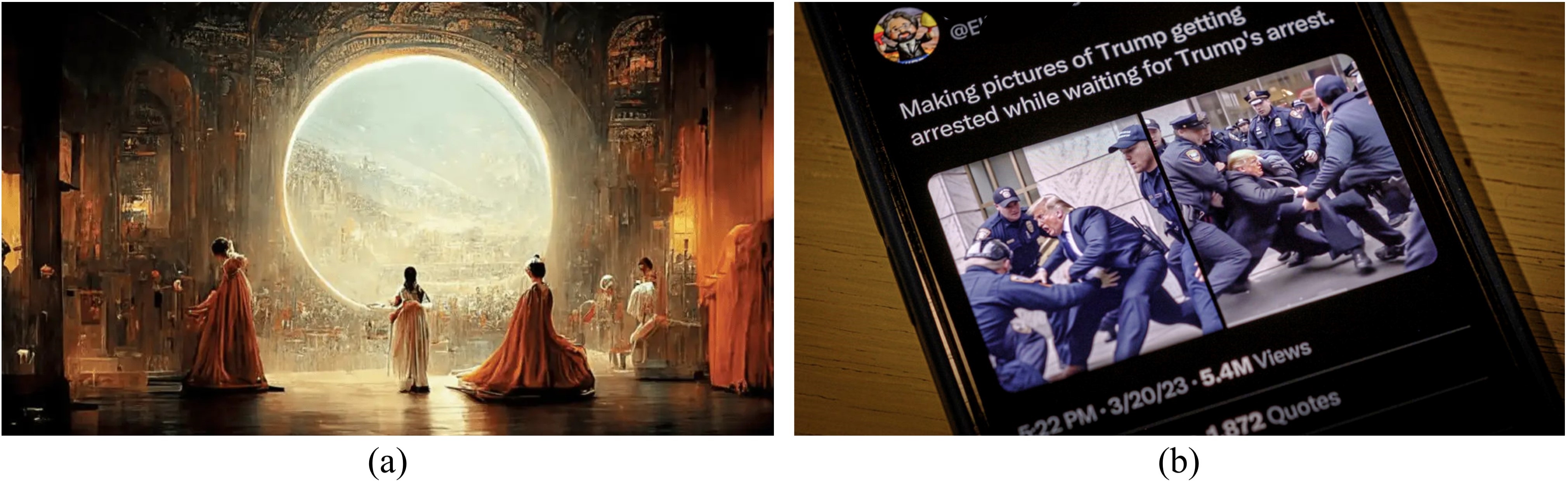}}
	\caption{AI-generated content may (a) inadvertently infringe on artistic copyrights or (b) maliciously propagate misinformation.}
	\label{fig:demo}
\end{figure}

\IEEEPARstart{T}{he} unyielding progress of deep learning has given rise to numerous prominent generative models, including \underline{G}enerative \underline{A}dversarial \underline{N}etworks (GANs) and \underline{D}iffusion \underline{M}odels (DMs). The photorealism and creativity exhibited by the images synthesized through these models have received increasing attention from various communities. In August of 2022, a DM-generated painting named Theater d'Opera Spatial (see Fig.~\ref{fig:demo} (a)) claimed the first prize at the Colorado State Fair's digital art competition, catapulting generative models into the spotlight. While generative models may serve as a source of inspiration for artists and designers or provide entertainment, there is a grave apprehension regarding their potential for malicious use in the generation and dissemination of misinformation (see Fig.~\ref{fig:demo} (b)). Under this circumstance, it is urgent to develop generalizable forensic algorithms, capable of distinguishing synthetic images from real ones.

\begin{figure*}[t]
	\centering
	\subfloat{
		\includegraphics[width = 0.98\textwidth]{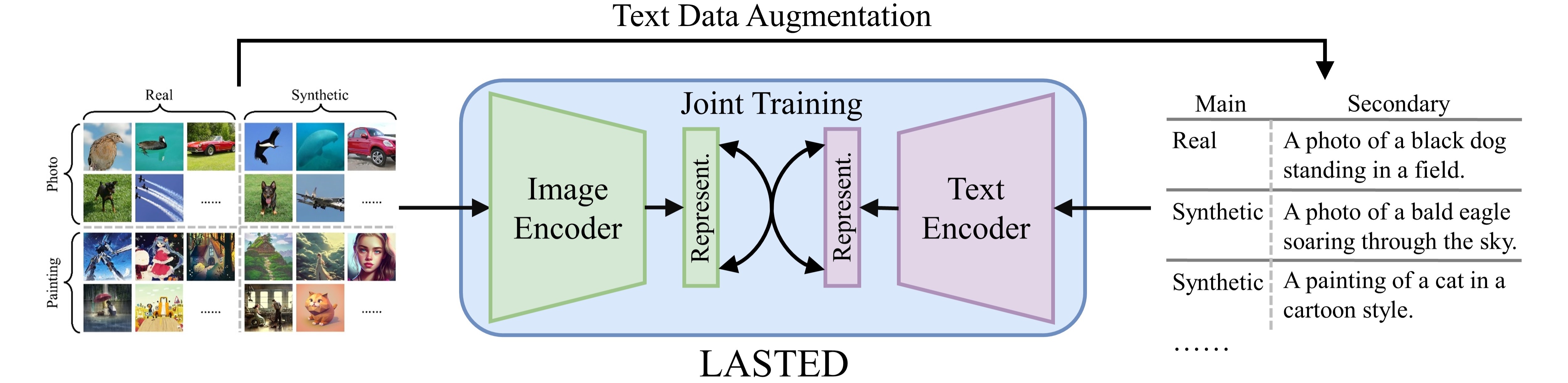}}
	\caption{Illustration of our proposed LASTED. The training images are first augmented with the carefully-designed textual labels, and then image/text encoders are jointly trained.}
	\label{fig:demo_lasted}
\end{figure*}

\begin{figure}[t!]
	\centering
	\subfloat{
		\includegraphics[width = 0.48\textwidth]{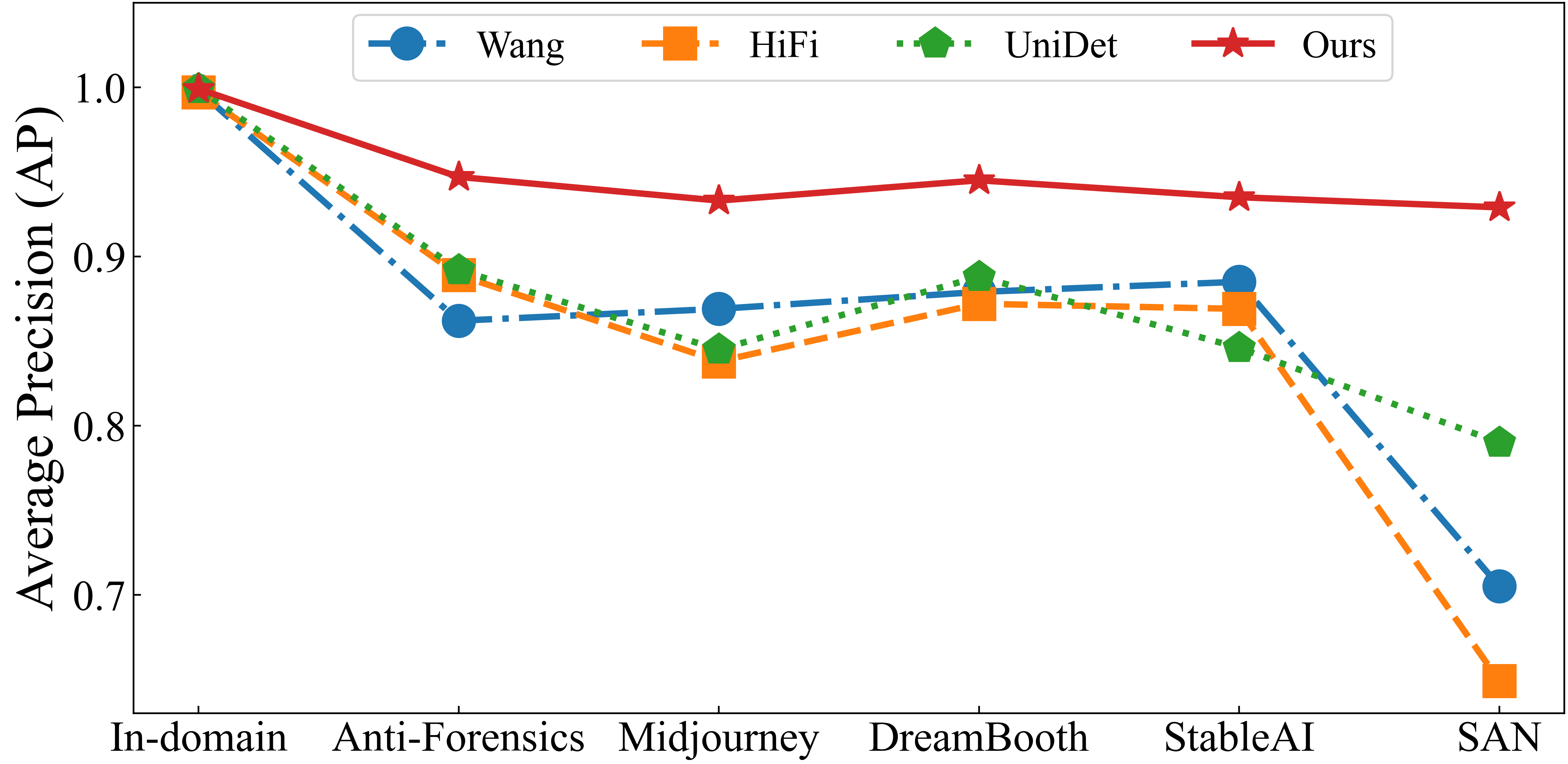}}
	\caption{Existing detectors (\eg, Wang \cite{easyspot2020}, HiFi \cite{hifi}, UniDet \cite{unidet}) suffer from severe performance degradation in out-of-domain scenarios (Midjourney \cite{midjourney}, DreamBooth \cite{dreambooth}, SAN \cite{san}).}
	\label{fig:head_fig}
\end{figure}

Several algorithms have been proposed to improve the generalizability of synthetic image detectors through model transferability \cite{deepfingerprint}, data adaptability \cite{unidet}, data augmentation \cite{easyspot2020, improving2023}, etc. Despite the relative success of these algorithms, their performance, especially the generalization capability, is still far from being adequate to cope with the increasing number of synthetic models. As can be seen in Fig.~\ref{fig:head_fig}, existing algorithms Wang \cite{easyspot2020}, HiFi \cite{hifi}, and UniDet \cite{unidet} fail to generalize to detect unseen synthesis approaches.

In this work, we aim to enhance the generalizability of the synthetic image detector by rethinking the training paradigm. From the training perspective, a large body of works~\cite{mae, clip, moco, circleloss} (and references therein) have shown that contrastive learning can enhance the general representation ability of neural networks. In particular, CLIP~\cite{clip} is a model developed by OpenAI that learns to associate images and text, enabling it to understand visual concepts in a more generalized way. It is based on language supervision and contrastive learning, and can even compete with supervised models in many tasks. Inspired by this, we propose a new synthetic image detection method: \underline{L}angu\underline{A}ge-guided \underline{S}yn\underline{T}h\underline{E}sis \underline{D}etection (LASTED), which utilizes an augmented language supervision to improve the image-domain forensic feature extraction. Noticing that the training data (synthetic or real images) usually do not accompany with textual information, we suggest to augment them with carefully-designed textual labels. We first assign main labels ``Real'' or ``Synthetic'' based on the images sources. Considering that a single main label may inadequately capture the diverse image distributions, we further design the secondary labels based on the image content. These secondary labels are generated by a pre-trained image captioning algorithm, \eg, ClipCap \cite{caption}, eliminating the need for manual annotation. Upon having the image-text pairs, our LASTED jointly trains an image encoder and a text encoder to predict the matched pairings of a batch of (image, text) examples under a contrastive learning framework. Essentially, the augmented textual labels provide learnable high-dimensional targets, which do not have to be composed of orthogonal one-hot vectors, thus making the semantic decoupling easier to be optimized. The contrastive training process of our LASTED is illustrated in Fig.~\ref{fig:demo_lasted}. After the training, we utilize linear probe to distill the learned knowledge and implement the specific synthetic detection.

Experiments show that our LASTED achieves much improved generalizability to unseen image generation models (see Fig.~\ref{fig:head_fig}). More specifically, LASTED delivers promising performance that far exceeds state-of-the-art competitors \cite{easyspot2020, color_robust2022, grag2021, lgrad, dire, hifi, unidet} on existing datasets ForenSynths \cite{easyspot2020} and DF$^3$ \cite{df3} with +5.5\% and +3.4\% average precision (AP), respectively. Further, as existing datasets \cite{easyspot2020, df3} primarily focus on synthetic images in \textit{photo} styles, we gather new datasets with \textit{painting} style. Clearly, paintings contain fantasy elements and lack camera traces, resulting in completely different forensic features (see Sec.~\ref{sec:exp:setting} and Fig.~\ref{fig:practical} for more details). These new datasets would allow for a more comprehensive evaluation of various detection algorithms. Contributions of this work can be summarized as:  
\begin{itemize}
	
	\item By incorporating carefully-designed textual labels, we devise LASTED for detecting synthetic images based on language supervision, capable of extracting highly discriminative forensic features from the joint visual-language space under a contrastive learning framework.      
	
	\item We collect synthetic image datasets featuring a painterly style to better emulate the real-world diversity.
	
	\item Experimental results demonstrate the superiority of the proposed LASTED compared with state-of-the-art methods \cite{easyspot2020, color_robust2022, grag2021, lgrad, dire, hifi, unidet} over four diverse datasets.
\end{itemize}

\begin{figure*}[th!]
	\centering
	\subfloat{
		\includegraphics[width = 0.98\textwidth]{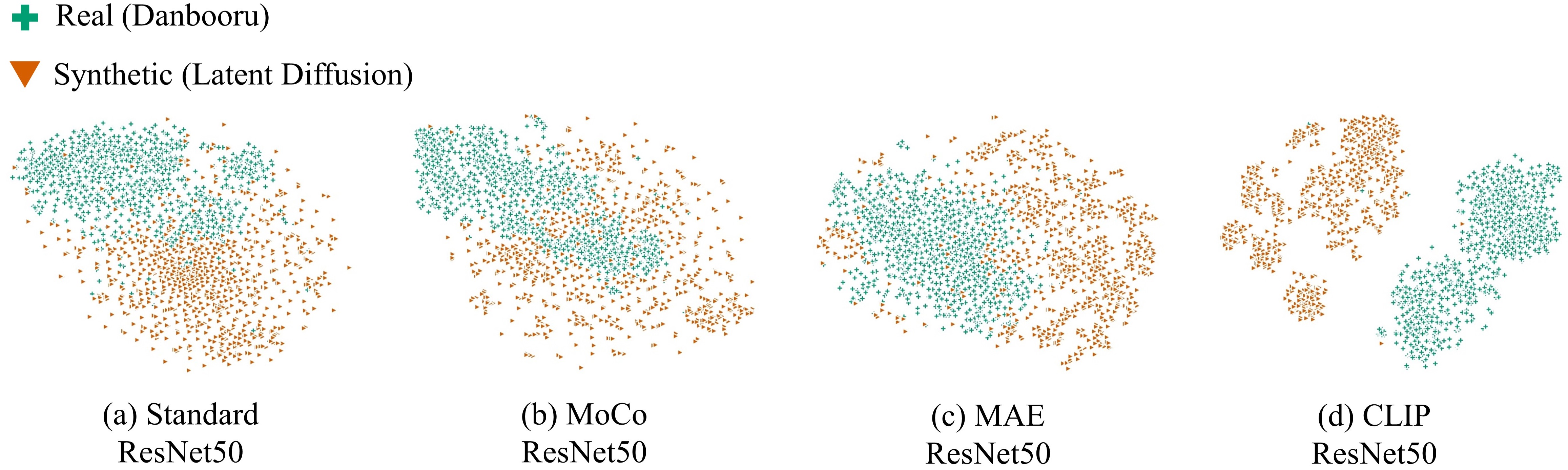}}
	\caption{Different training paradigms (MoCo \cite{moco}, MAE \cite{mae}, and CLIP \cite{clip}) lead to different generalizability. It is worth noting that all models use the same ResNet50 \cite{resnet2016} architecture and are trained on the natural dataset ImageNet \cite{deng2009imagenet} without fine-tuning. The testing is carried out on unseen real painting images (Danbooru \cite{danbooru2021}) and the ones synthesized by Latent Diffusion \cite{latent_diffusion}.}
	\label{fig:zeroshot_cmp}
\end{figure*}

\section{Related Works on Synthetic Image Detection}\label{sec:related_words}
In recent years, many detection algorithms \cite{odena2016deconvolution, marra2019incremental, chai2020makes, sha2022fake, easyspot2020, dzanic2020fourier, durall2020watch, zhang2019detecting, frank2020leveraging, color_robust2022, vcip2021, eccv2022fingerprintnet, iccv2021towards, icip2022fusing, icme2022improving, icml2021noiseinjection, ijcai2021, joi2021, lgrad} have been proposed to combat the potential malicious use of AI-generated images. Typically, these algorithms leverage the unique traces left by the image synthesis process, such as checkerboard \cite{odena2016deconvolution}, color \cite{color_robust2022}, and gradient \cite{lgrad}. Through frequency-domain analysis, Dzanic \etal \cite{dzanic2020fourier} and Durall \etal \cite{durall2020watch} showed that GAN-generated images deviate from real data in terms of spectral distribution. In this regard, some studies \cite{frequencyaware, detectingbyreal, thinktwice, towardsdiffdeepfake, intriguing} facilitated the extraction of synthetic artifacts by introducing frequency-aware attentional feature or learnable noise pattern in amplitude and phase spectra domains. Considering the difficulties of applying a detector to detect unknown types of image, some other approaches suggested enhancing the detector's generalization capability through the utilization of data augmentation \cite{easyspot2020, improving2023} and knowledge distillation \cite{unidet}. Although effective, \cite{dm_detection_2022, towardsdiffdeepfake} disclosed that detectors trained only on GAN-generated images cannot generalize well to detect DM-generated ones. Their results indicate that DM images are characterized by distinct artifacts from those of GAN images. In pursuit of stronger generalization, recent works \cite{cozzolino2024raising, cioni2024clip} explored lightweight fine-tuning strategies based on large pre-trained vision-language models (VLMs), demonstrating significant improvements in generalization to out-of-domain evaluation.

\section{LASTED for Synthetic Image Detection}\label{sec:method}
Our goal is to design a synthetic image detector, termed LASTED, that can generalize well to unseen data. This is achieved by improving the training paradigm with the assistance of language supervision. In the following, we first introduce the motivation of utilizing augmented language supervision, followed by the details on how to design the textual labels. Eventually, the detailed training procedures of our LASTED are given.

\subsection{Motivation}

The training paradigm based on contrastive learning has been proven to have strong generalizability and zero-shot transferability in image classification \cite{deng2009imagenet, moco, mae, clip}. Inspired by this, we attempt to involve the contrastive learning in the task of synthetic image detection. To explore which contrastive paradigm is the most suitable, we evaluate three widely adopted paradigms, namely, MoCo \cite{moco}, MAE \cite{mae}, and CLIP \cite{clip}, on their abilities to extract generalizable representations for discriminating unseen real (Danbooru \cite{danbooru2021}) and synthetic (Latent Diffusion \cite{latent_diffusion}) painting images. The extracted features, visualized by T-SNE \cite{tsne}, are shown in Fig.~\ref{fig:zeroshot_cmp}. Note that no fine-tuning is conducted. It can be seen that although these models are only trained on real natural photos (\eg, ImageNet \cite{deng2009imagenet} that is a large-scale dataset widely used for evaluating image classification models, containing millions of labeled images across thousands of categories), they can still extract highly discriminative representations from unseen real/synthetic painting images. In particular, the representations extracted by CLIP can distinguish painting images from Danbooru and Latent Diffusion in a much more satisfactory manner. We therefore speculate that although DM or GAN synthesized images have good visual realism, they can be still easily distinguished in the joint visual-language feature space.

We attribute the formation of such a highly discriminative feature space to the richness of information and multimodal representations. Specifically, textual descriptions provide a wealth of information that goes beyond fixed labels. This variability enables the detector to capture the discriminative properties of different contexts, thus improving its ability to distinguish whether similar context belongs to the real or synthetic category. On the other hand, the multimodal objective supervise the detector to identify not just the visual features of images but also how these features relate to textual descriptions. This supervision effectively shapes the feature space, leading to better separation and organization of representations. Now we are ready to design appropriate textual labels and conduct multimodal training based on them.

\begin{table}[t]
	\centering
	\scalebox{1.0}{
		\begin{tabular}{lll}
			\hline
			\hline
			\# & Main Labels & Secondary Labels\\
			\hline
			$\mathcal{R}_1$ & ``Real/Synthetic'' & \\
			$\mathcal{R}_2$ & ``Real/Synthetic'' & ``Photo/Painting'' \\
			$\mathcal{R}_3$ & ``Real/Synthetic'' & [ImageNet Class] \\
			$\mathcal{R}_4$ & ``Real/Synthetic'' & [Image Captioning] \\
			$\mathcal{R}_5$ & ``Real/Synthetic'' & ``A photo/painting of [Image Captioning]'' \\
			\hline
			$\mathcal{R}_6$ & \multicolumn{2}{l}{``A real/synthetic photo/painting of [Image Captioning]''} \\
			\hline
			\hline
	\end{tabular}}
	\caption{Different textual labeling strategies. $\mathcal{R}_5$ is adopted in our LASTED.}
	\label{tab:label}
\end{table}

\begin{figure*}[t!]
	\centering
	\subfloat{
		\includegraphics[width = 0.98\textwidth]{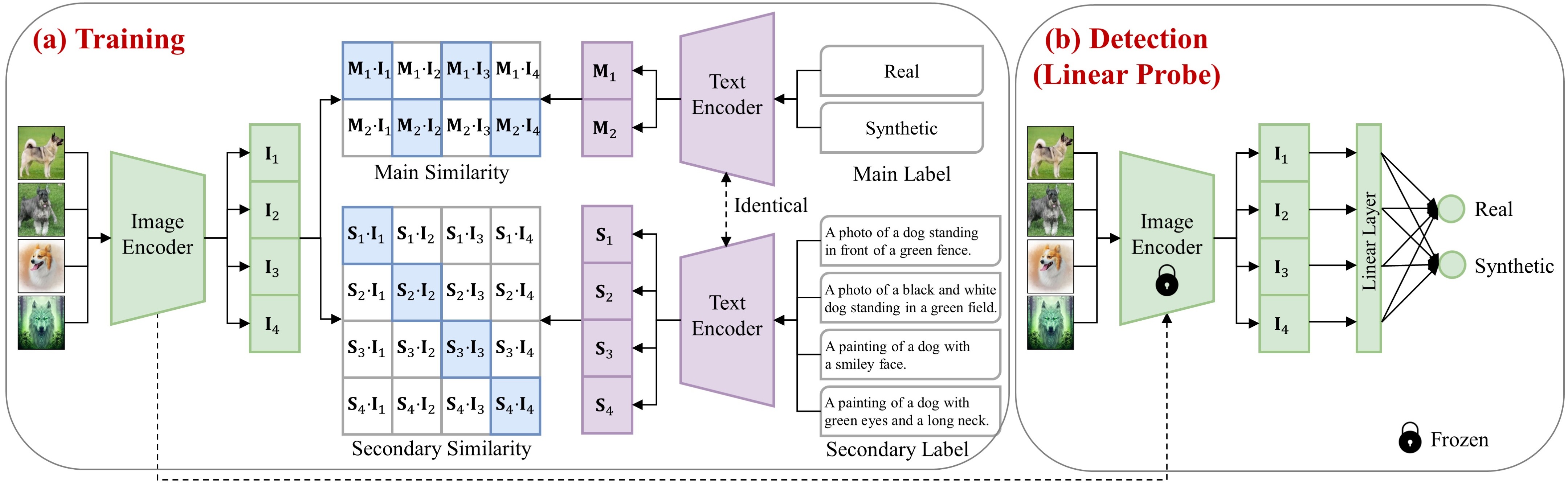}}
	\caption{Our proposed LASTED framework introduces a language-guided contrastive paradigm in its training process to better decouple the semantic information of the image and make the detector focus on the forensic signal. After the training is completed, the specific synthetic/real detection can be achieved by fine-tuning a lightweight linear probe.}
	\label{fig:framework}
\end{figure*}

\subsection{Augmenting with Textual Labels}\label{sec:method_label}

The first challenge for exploiting joint visual-language features is that the training images (real or synthetic) do not naturally accompany with textual information. It is hence crucial to design an appropriate text data augmentation strategy to specifically fit the synthetic image detection task. A naive method, to this end, is to manually associate a single word ``Real'' or ``Synthetic'' to each image in the training dataset (see $\mathcal{R}_1$ in Table~\ref{tab:label}), based on whether it is real or synthesized. However, we experimentally find that this simple labeling scheme leads to rather poor generalization in the synthetic image detection task. This is because there are different types of real (similarly synthetic) images, such as those captured by cameras (\eg, ImageNet \cite{deng2009imagenet}) and the ones drawn by humans through Photoshop or digital plates (\eg, Danbooru \cite{danbooru2021}). Clearly, these two types of real images have significantly different forensic features, \eg, the presence of camera traces in real photos and fictional elements in painting images, and hence, should not be mixed into the same category with a single ``Real'' label.

To solve this problem, we could append the secondary labels ``Photo'' and ``Painting'' to distinguish the aforementioned two types of images (see $\mathcal{R}_2$ in Table~\ref{tab:label}). In addition to the utilization of ``Photo/Painting'', we can also use the image semantics to further refine textual labels. Specifically, $\mathcal{R}_3$ employs a classifier trained on the ImageNet with 1000 categories to classify given images, and assign the predicted labels as the secondary labels for those images. Similarly, image captioning \cite{caption, li2020oscar} can be utilized to generate a sentence describing the content of the images, as given by $\mathcal{R}_4$. To more appropriately incorporate ``Photo/Painting'' into semantic information, we employ a template ``A photo/painting of [Image Captioning]'' to combine $\mathcal{R}_2$ and $\mathcal{R}_4$, resulting in $\mathcal{R}_5$, which eventually is used as the textual labels in the training process of our LASTED.   

Now, we are ready to introduce the training process of our LASTED using $\mathcal{R}_5$ as the textual labels.

\textit{Remark:} A natural idea is to directly combine the main labels and secondary labels, transforming $\mathcal{R}_5$ to $\mathcal{R}_6$. However, this labeling approach would significantly interfere with our core task of distinguishing between real and synthetic images, resulting in learning \textit{general} semantic features. Ablation studies to be presented in Sec.~\ref{sec:ablation} will show that $\mathcal{R}_6$ leads to a maximum 24.5\% decrease compared with $\mathcal{R}_5$ in the detection performance.

\subsection{Training Process of LASTED}
The training procedure of our LASTED is depicted in Fig.~\ref{fig:framework} (a), which mainly involves two encoders, namely, the image encoder $f_\Btheta$ and the text encoder $g_\Bphi$. Given a dataset $\mathcal{D}$ consisting of $N$ paired images and their augmented textual labels $\{(\mathbf{X}_i, \mathbf{Y}^m_i, \mathbf{Y}^s_i)\}_{i=1}^N$, the encoders first extract visual and textual representations $\mathbf{I}_i = f_\Btheta(\mathbf{X}_i)$, $\mathbf{M}_i = g_\Bphi(\mathbf{Y}^m_i)$, and $\mathbf{S}_i = g_\Bphi(\mathbf{Y}^s_i)$. Here, $\mathbf{Y}^m_i$ and $\mathbf{Y}^s_i$ are the main and secondary textual labels for the $i$th image $\mathbf{X}_i$, respectively. By performing dot product on visual and textual representations, we obtain similarity matrices for both main and secondary labels. The objective function is designed to maximize the cosine similarity of matched visual and textual pairings (see blue boxes in Fig.~\ref{fig:framework} (a)), while minimizing the unmatched ones (white boxes). Formally, the loss for main similarity matrix along the image axis can be expressed as:
\begin{equation}
	\mathcal{L}^m_I = \frac{1}{B}\sum_{i=1}^B -\log\frac{\exp{(\mathbf{I}_i \cdot \mathbf{M}_i / \tau)}}{\sum_{j \in \llbracket 1, C \rrbracket}\exp{(\mathbf{I}_i \cdot \mathbf{M}_j / \tau)}}.
\end{equation} Similarly, we can calculate the loss along the text axis by:
\begin{equation}
	\mathcal{L}^m_T = \frac{1}{C}\sum_{j=1}^C -\log \frac{\sum_{k \in \llbracket 1, B \rrbracket, \mathbf{M}_k=\mathbf{M}_j}\exp{(\mathbf{M}_k \cdot \mathbf{I}_k / \tau)}}{\sum_{i \in \llbracket 1, N \rrbracket}\exp{(\mathbf{M}_j \cdot \mathbf{I}_i / \tau)}},
\end{equation}
where $\tau$ is a learned temperature parameter \cite{clip}, and $C$ represents the number of distinct labels in the current batch having $B$ images. $\llbracket i,j \rrbracket$ denotes the integer interval from $i$ to $j$. Similarly, we can calculate the image-axis $\mathcal{L}^s_I$ and text-axis $\mathcal{L}^s_T$ for the secondary similarity matrix. It should be noted that as the similarity matrix may not always be square, for images with the same textual labels, we simply average their similarities. The overall loss then becomes 
\begin{equation}
	\mathcal{L} = \mathcal{L}^m_I + \mathcal{L}^m_T + \lambda(\mathcal{L}^s_I + \mathcal{L}^s_T),
\end{equation}
where $\lambda$ is the trade-off parameter between the main and secondary losses.

Regarding the network architectures, we adopt ResNet50x64 \cite{clip} and Text Transformer \cite{radford2019language} for the image encoder $f_\Btheta$ and text encoder $g_\Bphi$, respectively. It should also be noted that the selection of networks is flexible, as long as the extracted representations $\mathbf{I}$, $\mathbf{M}$, and $\mathbf{S}$ share the same dimensional feature space. The ablation studies on using different image encoders will be given in Sec.~\ref{sec:ablation}.

\subsection{Detection Process of LASTED}\label{sec:LASTED_testing}

Upon encoders training, we propose to implement the synthetic image detection through a linear probe (LP) approach, as illustrated in Fig.~\ref{fig:framework} (b). LP aims to distill the knowledge learned from encoders, so as to be more general in detecting unseen synthetic images \cite{gu2022open, unidet}. Specifically, we extend an additional linear layer $l_\tau$ on top of the well-trained image encoder $f_\theta$ and supervise it using binary cross-entropy loss:
\begin{equation}
	\mathcal{L}_{ce} = -\sum_{i \in \mathcal{D}^S}\log(l_\tau(f_\theta(\mathbf{X}_i))) - \sum_{j \in \mathcal{D}^R}\log(1-l_\tau(f_\theta(\mathbf{X}_j))),
\end{equation}
where $\mathcal{D}^R$ and $\mathcal{D}^S$ denote the indices of real and synthetic subset of dataset $\mathcal{D}$. It should be noted that the image encoder $f_\theta$ remains frozen and does not require further fine-tuning, mainly because it has already been supervised to have a feature space with good discriminability during the previous contrastive learning. Since the linear layer $l_\tau$ contains very few parameters (\eg, $\tau \in \mathbb{R}^{1024}$), the fine-tuning process can be completed quickly.

Alternatively, we can also implement the detection process through the nearest neighbor (NN) matching. Specifically, we first extract text representations $\mathbf{M}_r$ and $\mathbf{M}_s$ for ``Real'' and ``Synthetic'' categories, respectively. Given a queried image $\mathbf{X}_t$, we then calculate the cosine similarity between its representation $\mathbf{I}_t = f_\theta(\mathbf{X}_t)$ and $\mathbf{M}_r$ or $\mathbf{M}_s$. The probability $p$ of $\mathbf{X}_t$ belonging to the ``Synthetic'' category can be derived through a SoftMax function:
\begin{equation}
	p = \frac{\exp(\mathbf{I}_t \cdot \mathbf{M}_s)}{\exp(\mathbf{I}_t \cdot \mathbf{M}_r) + \exp(\mathbf{I}_t \cdot \mathbf{M}_s)}.
\end{equation}
As will be clear soon, we empirically find that the NN approach performs worse than the LP approach. This may be because ``Real'' and ``Synthetic'' labels are affected by the secondary labels, resulting in insufficient discriminability.

\begin{table}[t!]
	\centering
	\scalebox{0.85}{
		\centering
		\begin{tabular}{lcccc}
			\hline
			\hline
			\multirow{2}{*}{Sets} & \multicolumn{2}{c}{Training} & \multicolumn{2}{c}{Testing}\\
			\cline{2-5} & Real & Synthetic & Real & Synthetic \\
			\cline{1-5}
			\multirow{2}{*}{$\mathcal{T}_{gan}$ \cite{easyspot2020}} & \multirow{4}{*}{LSUN} & \multirow{4}{*}{ProGAN} & LSUN, ImageNet & \multirow{2}{*}{10 GANs} \\
			& & & CelebA, GTA & \\
			\cline{1-1} \cline{4-5} \multirow{2}{*}{$\mathcal{T}_{df}$ \cite{df3}} & & & \multirow{2}{*}{FFHQ} & 3 GANs, 2 DMs \\
			& & & & Transformers \\
			\hline
			\multirow{2}{*}{$\mathcal{T}_{fuse}$} & \multirow{6}{*}{\shortstack{LSUN \\ Danbooru}} & \multirow{6}{*}{\shortstack{ProGAN \\ SD}} & ImageNet, VISION, & 3 GANs \\
			& & & Artist, Danbooru & 4 DMs \\
			\cline{1-1} \cline{4-5}
			\multirow{4}{*}{$\mathcal{T}_{wild}$} & & & \multirow{4}{*}{Artist} & DreamBooth \\
			& & & & Midjourney \\
			& & & & NightCafe \\
			& & & & YiJian, SAI \\
			\hline
			\hline
		\end{tabular}
	}
	\caption{Lists of training and testing datasets.}
	\label{tab:datasets}
\end{table}

\begin{table*}[t!]
	\centering
	\scalebox{0.98}{
		\begin{tabular}{lcccccccccccc}
			\hline
			\hline
			Methods & Variants & StyleGAN & BigGAN & CycleGAN & StarGAN & GauGAN & CRN & IMLE & SITD & SAN & Deepfake & Mean \\
			\hline
			\multirow{2}{*}{Wang \cite{easyspot2020}} & B+J (0.5) & .985 & .882 & .968 & .954 & .981 & .989 & .995 & .927 & .639 & .663 & .898 \\
			& B+J (0.1) & .996 & .845 & .935 & .982 & .895 & .982 & .984 & .972 & .705 & \textbf{.890} & .919 \\
			\cline{1-2}
			Grag \cite{grag2021} & LP & .963 & .891 & .924 & .988 & .912 & .927 & .927 & .764 & .520 & .519 & .834 \\
			\cline{1-2}
			CR \cite{color_robust2022} & LP & .986 & .881 & .957 & .939 & .990 & .993 & .998 & .799 & .657 & .716 & .892 \\
			\cline{1-2}
			\multirow{3}{*}{LGrad \cite{lgrad}} & 1class & .990 & .844 & .891 & .998 & .772 & .770 & .767 & .435 & .480 & .828 & .777 \\
			& 2class & .992 & .888 & .923 & .999 & .773 & .567 & .753 & .435 & .465 & .722 & .752 \\
			& 4class & .996 & .873 & .919 & .999 & .739 & .583 & .658 & .396 & .442 & .755 & .736 \\
			\cline{1-2} DIRE \cite{dire} & LP & .985 & .942 & .917 & .919 & .960 & .933 & .891 & .835 & .696 & .780 & .886 \\
			\cline{1-2} HiFi \cite{hifi} & LP & .964 & .921 & .910 & .958 & .955 & .826 & .898 & .902 & .649 & .738 & .872 \\
			\cline{1-2}
			UniDet \cite{unidet} & LP & .972 & \textbf{.996} & .995 & .996 & \textbf{.999} & .967 & .990 & .613 & .790 & .825 & .914  \\
			\hline
			\multirow{2}{*}{LASTED} & NN & .918 & .912 & .965 & .994 & .924 & .938 & .938 & .970 & .905 & .809 & .927  \\
			& LP & \textbf{.999} & .983 & \textbf{.997} & \textbf{.999} & .986 & \textbf{.999} & \textbf{.999} & \textbf{.973} & \textbf{.929} & .873 & \textbf{.974}  \\
			\hline
			\hline
		\end{tabular}
	}
	\caption{Detection results on $\mathcal{T}_{gan}$ by using AP as a criterion.}
	\label{tab:cmp_progan}
\end{table*}

\section{Experiments}\label{sec:experiments}
\subsection{Settings}\label{sec:exp:setting}
\textbf{Dataset.} We establish four challenging datasets to comprehensively evaluate the performance of our proposed LASTED. A brief summary is given in Table.~\ref{tab:datasets}.

	$\mathcal{T}_{gan}$ \cite{easyspot2020}: Following \cite{easyspot2020, color_robust2022, grag2021, lgrad, unidet}, we adopt the ForenSynths \cite{easyspot2020} dataset (named $\mathcal{T}_{gan}$), comprised of 11 subsets generated by 11 synthetic approaches, including ProGAN \cite{progan}, StyleGAN \cite{stylegan}, BigGAN \cite{biggan}, CycleGAN \cite{cyclegan}, StarGAN \cite{stargan}, GauGAN \cite{gaugan}, CRN \cite{crn}, IMLE \cite{imle}, SITD \cite{sitd}, SAN \cite{san}, and Deepfake \cite{deepfake}. To assess how well the synthetic image detectors generalize to unseen images, only the ProGAN subset is utilized for training, while the remaining 10 subsets are used for testing. 
		
	$\mathcal{T}_{df}$ \cite{df3}: Ju \etal \cite{df3} stated that the $\mathcal{T}_{gan}$ dataset cannot reflect the real-world situations, and they instead created the $\mathcal{T}_{df}$ dataset by combining 6 advanced generative models and 5 anti-forensic operations. Specifically, $\mathcal{T}_{df}$ involves 46,400 synthetic images generated by 3 GANs (3DGAN \cite{3dgan}, StyleGAN2 \cite{styelgan2}, StyleGAN3 \cite{stylegan3}), 2 DMs (LSGM \cite{lsgm}, Latent Diffusion \cite{latent_diffusion}) and 1 Transformers \cite{esser2021taming}. To better simulate real-world situations, five post-processing operations are applied to counter forensic algorithms, including commonly used post-processing (\eg, compression and blurring), blending, anti-forensics (\eg, CW attack \cite{cw}), multi-image compression, and a mixture of the above operations. Note that the $\mathcal{T}_{df}$ dataset is only used for testing, while training is performed on $\mathcal{T}_{gan}$.

	$\mathcal{T}_{fuse}$: Considering that no painting images are included in $\mathcal{T}_{gan}$ and $\mathcal{T}_{df}$, we additionally form a fused dataset $\mathcal{T}_{fuse}$ by including 4 categories of data. Specifically, real and synthetic photos are generated from the LSUN \cite{LSUN} and ProGAN \cite{progan}, respectively. Furthermore, the real and synthetic paintings are sourced from the Danbooru \cite{danbooru2021} and the Stable Diffusion (SD) \cite{novelai, stablediffusion_v15, lexica}. The image synthesis models ProGAN and SD here are deliberately trained on LSUN and Danbooru, respectively, forcing the detector to learn more discriminative representations from visually similar real and synthetic images. Each category contains 200K images, among which 1\% images are split as validation data. For testing, $\mathcal{T}_{fuse}$ adopts seven representative image synthesis models to generate cross-domain synthetic images, including 3 GANs (BigGAN \cite{biggan}, GauGAN \cite{gaugan}, and StyleGAN \cite{stylegan}) and 4 DMs (DALLE \cite{dalle2}, GLIDE \cite{glide}, Guided Diffusion \cite{guided_diffusion}, and Latent Diffusion \cite{latent_diffusion}). Also, we randomly sample images from 4 real datasets ImageNet \cite{deng2009imagenet}, VISION \cite{vision2017}, Danbooru \cite{danbooru2021}, and Artist \cite{artstation, behance}, where balanced sampling is adopted.
	
\begin{figure}[t!]
	\centering
	\subfloat{
		\includegraphics[width = 0.48\textwidth]{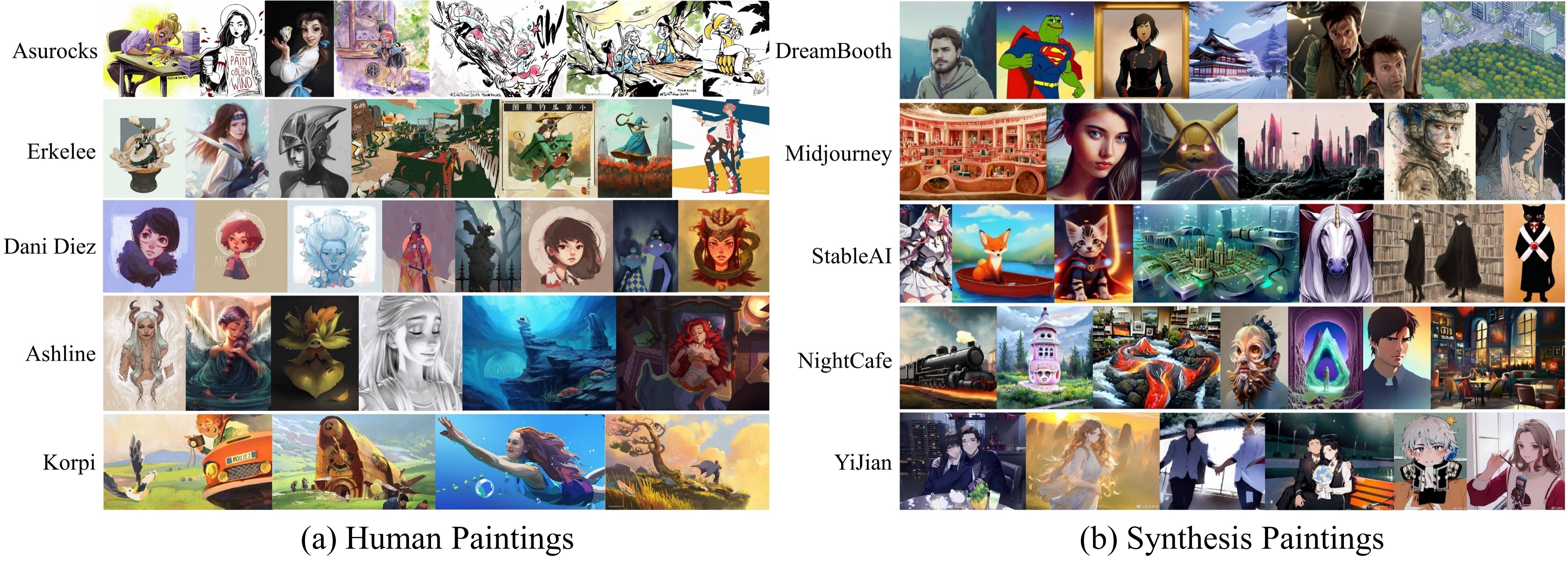}}
	\caption{(a) Each row presents paintings drawn by different artists from \cite{artstation, behance}; (b) Each row shows DM-generated images from different online platforms \cite{dreambooth, midjourney, nightcafe, stableai, yijian}.}
	\label{fig:practical}
\end{figure}

	$\mathcal{T}_{wild}$: In addition to $\mathcal{T}_{fuse}$, we also form a more challenging testing dataset $\mathcal{T}_{wild}$ by collecting images from mainstream sharing platforms. Real-world synthetic images shared and disseminated by users in sharing platforms exhibit much higher quality, compared to randomly generated ones using pre-trained GANs or DMs. Images from $\mathcal{T}_{wild}$ could better reflect situations in reality. We gather a total of 4K images from DreamBooth \cite{dreambooth}, Midjourney \cite{midjourney}, NightCafe \cite{nightcafe}, StableAI \cite{stableai}, and YiJian \cite{yijian}. Additionally, we obtain 2,229 real painting images drawn by 63 artists from open-source sharing platforms \cite{artstation, behance}. A preview of these images is available in Fig.~\ref{fig:practical}.

\begin{table*}[t!]
	\centering
	\scalebox{0.98}{
		\begin{tabular}{lcccccccc}
			\hline
			\hline
			Methods & Variants & Unprocessed & Post-processing & Blending & Anti-forensics & Multi-compression & Mixed & Mean \\
			\hline
			\multirow{2}{*}{Wang \cite{easyspot2020}} & B+J (0.5) & .878 & .930 & .913 & .862 & .822 & .961 & .894 \\
			& B+J (0.1) & .957 & .926 & .919 & .926 & .749 & .942 & .903 \\
			\cline{1-2}
			Grag \cite{grag2021} & LP & .920 & .859 & .911 & .893 & .786 & .918 & .881 \\
			\cline{1-2}
			CR \cite{color_robust2022} & LP & .941 & .922 & .920 & .851 & .817 & .943 & .899 \\
			\cline{1-2}
			\multirow{3}{*}{LGrad \cite{lgrad}} & 1class & .899 & .816 & .872 & .884 & .852 & .838 & .860 \\
			& 2class & .876 & .831 & .842 & .879 & .860 & .844 & .855 \\
			& 4class & .914 & .892 & .908 & .850 & .826 & .907 & .883 \\
			\cline{1-2} DIRE \cite{dire} & LP & .928 & .915 & .892 & .868 & .854 & .924 & .897 \\
			\cline{1-2} HiFi \cite{hifi} & LP & .959 & .932 & .901 & .889 & .867 & .950 & .916 \\
			\cline{1-2}
			UniDet \cite{unidet} & LP & .965 & .943 & .915 & .892 & .855 & .936 & .918 \\
			\hline
			\multirow{2}{*}{LASTED} & NN & .970 & .957 & .925 & .927 & .886 & .954 & .937 \\
			& LP & \textbf{.981} & \textbf{.981} & \textbf{.932} & \textbf{.947} & \textbf{.902} & \textbf{.969} & \textbf{.952} \\
			\hline
			\hline
		\end{tabular}
	}
	\caption{Detection results on $\mathcal{T}_{df}$ by using AP as a criterion.}
	\label{tab:cmp_df}
\end{table*}

\textbf{Competitors and Evaluation Metrics.} The following state-of-the-art synthetic image detectors Wang \cite{easyspot2020}, CR \cite{color_robust2022}, Grag \cite{grag2021}, LGrad \cite{lgrad}, DIRE \cite{dire}, HiFi \cite{hifi} and UniDet \cite{unidet} are selected as comparative methods. In particular, for detectors with different implementations, we categorize them as distinct variants to enable a more detailed comparison. For example, Wang \cite{easyspot2020} includes two variants with self-augmentation strategies using probabilities of 10\% and 50\% respectively. Their official models (trained on $\mathcal{T}_{gan}$ and $\mathcal{T}_{df}$) can be obtained from the links in their papers. To ensure the fairness, we also \textit{retrain} all the competitors on $\mathcal{T}_{fuse}$ and $\mathcal{T}_{wild}$, in addition to directly using their released versions. Following their tradition, we use the average precision (AP) for evaluating the detection performance.

\begin{table*}[t!]
	\centering
	\scalebox{0.88}{
		\begin{tabular}{lcccccccccccccccc}
			\hline
			\hline
			\multirow{3}{*}{Methods} & \multicolumn{7}{c}{ImageNet \& VISION} & \multicolumn{7}{c}{Danbooru \& Artist} & \multirow{3}{*}{Mix} & \multirow{3}{*}{Mean} \\
			\cline{2-15} & \multicolumn{3}{c}{GANs} & \multicolumn{4}{c}{DMs} & \multicolumn{3}{c}{GANs} & \multicolumn{4}{c}{DMs} & & \\
			& Big & Gau & Style & DALLE & GLIDE & Guided & Latent & Big & Gau & Style & DALLE & GLIDE & Guided & Latent & & \\
			\hline
			Wang \cite{easyspot2020} & .837 & .853 & .738 & .859 & .566 & .600 & .622 & .882 & .953 & .995 & .918 & .647 & .604 & .679 & .558 & .754 \\
			Grag \cite{grag2021} & .825 & .872 & .886 & .844 & .646 & .635 & .591 & .830 & .964 & .944 & .927 & .686 & .638 & .625 & .534 & .763 \\
			CR \cite{color_robust2022} & .848 & .854 & .804 & .873 & .693 & .706 & .538 & .827 & .919 & .967 & .921 & .519 & .717 & .631 & .593 & .761 \\
			LGrad \cite{lgrad} & .818 & .894 & .854 & .823 & .619 & .668 & .543 & .810 & .960 & .979 & .925 & .655 & .695 & .732 & .779 & .784 \\
			DIRE \cite{dire} & .887 & .890 & .837 & .864 & .648 & .600 & .641 & .873 & .929 & .890 & .916 & .820 & .665 & .719 & .797 & .798 \\
			HiFi \cite{hifi} & .865 & .842 & .869 & .820 & .709 & .729 & .734 & .877 & .965 & .943 & .908 & .756 & .610 & .681 & .829 & .809 \\
			UniDet \cite{unidet} & .844 & .897 & .836 & .788 & .648 & .673 & .704 & .881 & .992 & .857 & .915 & .774 & .705 & .692 & .839 & .803 \\
			\hline
			LASTED (NN) & .935 & .888 & .953 & .845 & .843 & .804 & .765 & .966 & .951 & .967 & .938 & .826 & .803 & .875 & .892 & .883 \\
			LASTED (LP) & \textbf{.979} & \textbf{.931} & \textbf{.985} & \textbf{.891} & \textbf{.861} & \textbf{.830} & \textbf{.782} & \textbf{.992} & \textbf{.985} & \textbf{.998} & \textbf{.961} & \textbf{.869} & \textbf{.805} & \textbf{.900} & \textbf{.920} & \textbf{.913} \\
			\hline
			\hline
		\end{tabular}
	}
	\caption{Detection results on open dataset $\mathcal{T}_{fuse}$ by using AP as a criterion.}
	\label{tab:cmp_open}
\end{table*}

\textbf{Implementation Details.} We implement our method using the PyTorch deep learning framework, where the Adam \cite{kingma2014adam} with default parameters is adopted as the optimizer. The learning rate is initialized to 1e-4 and halved if the validation accuracy fails to increase for 2 epochs until the convergence. The image encoder $f_{\bm{\theta}}$ is pre-trained on the ImageNet dataset, while the text encoder $g_{\bm{\phi}}$ utilizes the weight provided by CLIP. In the training/testing processes, all the input images are randomly/center cropped into 448$\times$448 patches. Image-domain augmentation, including compression, blurring and scaling, has been applied with 50\% probability, which was similarly adopted in \cite{easyspot2020, color_robust2022}. The trade-off parameter $\lambda$ is empirically set to 0.1. The batch size is set to 48 and the training is performed on 4 NVIDIA A100 GPU 40GB.

\begin{table*}[t!]
	\centering
	\scalebox{1.15}{
		\begin{tabular}{lccccccc}
			\hline
			\hline
			Methods & DreamBooth & MidjourneyV4 & MidjourneyV5 & NightCafe & StableAI & YiJian & Mean \\
			\hline
			Wang \cite{easyspot2020} & .879 & .875 & .869 & .893 & .885 & .828 & .871 \\
			Grag \cite{grag2021} & .902 & .913 & .881 & .920 & .880 & .842 & .890 \\
			CR \cite{color_robust2022} & .842 & .862 & .835 & .875 & .847 & .672 & .822 \\
			LGrad \cite{lgrad} & .854 & .903 & .858 & .793 & .888 & .836 & .855 \\
			DIRE \cite{dire} & .845 & .829 & .822 & .830 & .856 & .833 & .836 \\
			HiFi \cite{hifi} & .872 & .851 & .838 & .862 & .869 & .891 & .864 \\
			UniDet \cite{unidet} & .888 & .844 & .845 & .831 & .846 & .872 & .854 \\
			\hline
			LASTED (NN) & .903 & .897 & .885 & .912 & .899 & .922 & .903 \\
			LASTED (LP) & \textbf{.945} & \textbf{.932} & \textbf{.933} & \textbf{.966} & \textbf{.935} & \textbf{.937} & \textbf{.941} \\
			\hline
			\hline
		\end{tabular}
	}
	\caption{Detection results on practical dataset $\mathcal{T}_{wild}$ by using AP as a criterion.}
	\label{tab:cmp_prac}
\end{table*}

\subsection{Evaluation on Dataset $\mathcal{T}_{gan}$}\label{sec:forensynths}

The comparative results on $\mathcal{T}_{gan}$ are presented in Table~\ref{tab:cmp_progan}. In general, all considered methods \cite{easyspot2020, color_robust2022, grag2021, lgrad, unidet} exhibit desirable generalization capabilities to other GAN-generated images, achieving high detection precision (over .90 AP) on subsets StyleGAN, CycleGAN, StarGAN, and GauGAN. This phenomenon is not surprising, as many different GANs and GAN-like generative models tend to leave similar traces. However, the detection performance of the existing methods on SAN and Deepfake becomes much inferior, \eg UniDet only achieves .790 and .825 AP, respectively, demonstrating their inadequate generalizability. As can be seen in the last two rows of Table~\ref{tab:cmp_progan}, our LASTED (NN and LP) offers satisfactory detection performance for all the 10 testing cases, specifically with high AP of .929 on SAN. In summary, our LASTED surpasses the best result among competitors by a substantial margin, achieving a remarkable .974 AP. This validates that our LASTED could learn more generalizable representations for the GAN-synthesized image detection.

\subsection{Evaluation on Dataset $\mathcal{T}_{df}$}\label{sec:df}
Results on Table~\ref{tab:cmp_df} show that anti-forensic operations can lead to detection performance degradation. For example, the performance of the UniDet \cite{unidet} drops from .965 to .892 and .855 in anti-forensics and multi-compression cases respectively. Our LASTED can still achieve good anti-interference performance, with an average AP of .952, surpassing the second-place UniDet's .918 by +3.4\%.

\begin{table*}[t!]
	\centering
	\scalebox{1.2}{
		\begin{tabular}{lcclcccccccc}
			\hline
			\hline
			\multirow{3}{*}{\#} & \multirow{3}{*}{Training Paradigm} & \multirow{3}{*}{Architecture} & \multirow{3}{*}{\shortstack[l]{Label Choice \\ in Table~\ref{tab:label}}} & \multicolumn{6}{c}{Testing Scenarios} & \multicolumn{2}{c}{\multirow{2}{*}{Mean}} \\
			\cline{5-10} & & & & \multicolumn{2}{c}{$\mathcal{T}_{gan}$} & \multicolumn{2}{c}{$\mathcal{T}_{fuse}$} & \multicolumn{2}{c}{$\mathcal{T}_{wild}$} & & \\
			& & & & NN & LP & NN & LP & NN & LP & NN & LP \\
			\hline
			\#1 & Classification & RN50x64 & - & - & .907 & - & .752 & - & .868 & - & .842 \\
			\hline
			\#2 & Img-Img Contrastive & RN50x64 & - & - & .875 & - & .724 & - & .810 & - & .803 \\
			\hline
			\#3 & Img-Txt w/o Eq. (1) & \multirow{2}{*}{RN50x64} & \multirow{2}{*}{$\mathcal{R}_5$ (ClipCap)} & .840 & .886 & .806 & .838 & .782 & .819 & .809 & .848 \\
			\#4 & Img-Txt w/o Eq. (2) & & & .851 & .894 & .820 & .823 & .775 & .854 & .815 & .857 \\
			\hline
			\#5 & \multirow{11}{*}{Img-Txt Contrastive} & \multirow{8}{*}{RN50x64} & $\mathcal{R}_1$ & .914 & .926 & .813 & .844 & .863 & .903 & .863 & .891 \\
			\#6 & & & $\mathcal{R}_2$ & .912 & .938 & .815 & .863 & .852 & .900 & .860 & .900 \\
			\#7 & & & $\mathcal{R}_3$ (1K) & .920 & .941 & .824 & .857 & .864 & .913 & .869 & .904 \\
			\#8 & & & $\mathcal{R}_3$ (21K) & .918 & .948 & .810 & .872 & .849 & .916 & .859 & .912 \\
			\#9 & & & $\mathcal{R}_4$ (Oscar) & .924 & .954 & .874 & .894 & .885 & .927 & .894 & .925 \\
			\#10 & & & $\mathcal{R}_4$ (ClipCap) & .921 & .967 & .867 & .900 & .893 & .934 & .894 & .934 \\
			\#11 & & & $\mathcal{R}_5$ (ClipCap) & \textbf{.927} & \textbf{.974} & \textbf{.883} & \textbf{.913} & \textbf{.903} & \textbf{.941} & \textbf{.904} & \textbf{.943} \\
			\#12 & & & $\mathcal{R}_6$ (ClipCap) & .708 & .847 & .665 & .849 & .605 & .873 & .659 & .856 \\
			\cline{3-12}
			\#13 & & ConvNeXt & $\mathcal{R}_5$ (ClipCap) & .922 & .959 & .855 & .874 & .889 & .912 & .889 & .915 \\
			\#14 & & ViT & $\mathcal{R}_5$ (ClipCap) & .903 & .935 & .821 & .855 & .893 & .904 & .872 & .898  \\
			\#15 & & MiT & $\mathcal{R}_5$ (ClipCap) & .905 & .913 & .834 & .854 & .871 & .890 & .870 & .886  \\
			\hline
			\hline
		\end{tabular}
	}
	\caption{Ablation studies regarding the training paradigm, architecture, and label representation. AP is adopted as the criterion. ``-'' means non-applicability. \#11 variant is the final combination used to implement LASTED.}
	\label{tab:ablation}
\end{table*}

\subsection{Evaluation on Dataset $\mathcal{T}_{fuse}$}\label{sec:openset}

The detection results on $\mathcal{T}_{fuse}$ of competing methods are tabulated in Table~\ref{tab:cmp_open}. It can be seen that the detection performance of Wang \cite{easyspot2020}, Grag \cite{grag2021}, and CR \cite{color_robust2022} is relatively poor; their average APs are around .50$\sim$.70 when detecting Guided and Latent DMs from real images, indicating that their models cannot extract discriminative representations in open scenarios. In addition, UniDet \cite{unidet} performs well in distinguishing real images from GAN synthetic images, which can be reflected by the .992 AP when detecting GauGAN from Danbooru and Artist. However, it still struggles to generalize well to other detection scenarios. For instance, APs are only .648 when distinguishing GLIDE from ImageNet and VISION, and .705 when classifying Guided Diffusion from Danbooru and Artist. In contrast, aided by the language supervision, our proposed LASTED exhibits desirable detection performance for most considered cases. Noted that in Table~\ref{tab:cmp_open}, we include a ``mix'' column to indicate the integration and simultaneous inference of all classes from $\mathcal{T}_{fuse}$, in contrast to the previous columns where only a single class is inferred each time. From the result, it can be concluded that LASTED possesses superior generalizability, allowing it to extract highly discriminative representations in multiple cross-domain testing datasets. Overall, our LASTED achieves an average AP of .913, significantly outperforming the second-ranked competitor by +10.4\%.

\subsection{Evaluation on Dataset $\mathcal{T}_{wild}$}\label{sec:practical}

Let us now evaluate the synthetic image detection performance on a more practical dataset $\mathcal{T}_{wild}$. As can be observed from Table~\ref{tab:cmp_prac}, existing algorithms generally achieve commendable performance. For instance, Grag \cite{grag2021} attains .902 and .913 AP in the DreamBooth and MidjourneyV4, respectively. Nevertheless, the classification-based training algorithms \cite{easyspot2020, grag2021, color_robust2022, lgrad, dire, hifi, unidet} posses rather limited generalizability, leading to inferior detection performance. In contrast, our language-guided contrastive paradigm empowers the model to learn more generalizable image representations, achieving an average of .941 AP, which outperforms the second place by +5.1\%.

\begin{table}[t!]
	\centering
	\scalebox{1.2}{
		\begin{tabular}{lccccc}
			\hline
			\hline
			\multirow{2}{*}{Datasets} & \multicolumn{4}{c}{Zero-shot Detection} & \multirow{2}{*}{LASTED} \\
			& Std. & MoCo & MAE & CLIP & \\
			\hline
			$\mathcal{T}_{gan}$ & .604 & .716 & .769 & .808 & .927 \\
			$\mathcal{T}_{df}$ & .547 & .676 & .723 & .789 & .937 \\
			$\mathcal{T}_{fuse}$ & .533 & .544 & .651 & .750 & .883 \\
			$\mathcal{T}_{wild}$ & .515 & .564 & .658 & .771 & .903 \\
			\hline
			\hline
		\end{tabular}
	}
	\caption{Zero-shot detection (in AP) by using standard, MoCo \cite{moco}, MAE \cite{mae}, and CLIP \cite{clip} paradigms. Our LASTED is provided for reference.}
	\label{tab:zero_shot}
\end{table}

\subsection{Ablation Studies}\label{sec:ablation}
In this subsection, we conduct the ablation studies of LASTED by analyzing how the training paradigm, language supervision, network architecture, and the loss parameter contribute to the eventual detection performance. The main comparative results over $\mathcal{T}_{gan}$, $\mathcal{T}_{fuse}$, and $\mathcal{T}_{wild}$ datasets are given in Table~\ref{tab:ablation}.

\begin{figure}[t!]
	\centering
	\subfloat{
		\includegraphics[width = 0.45\textwidth]{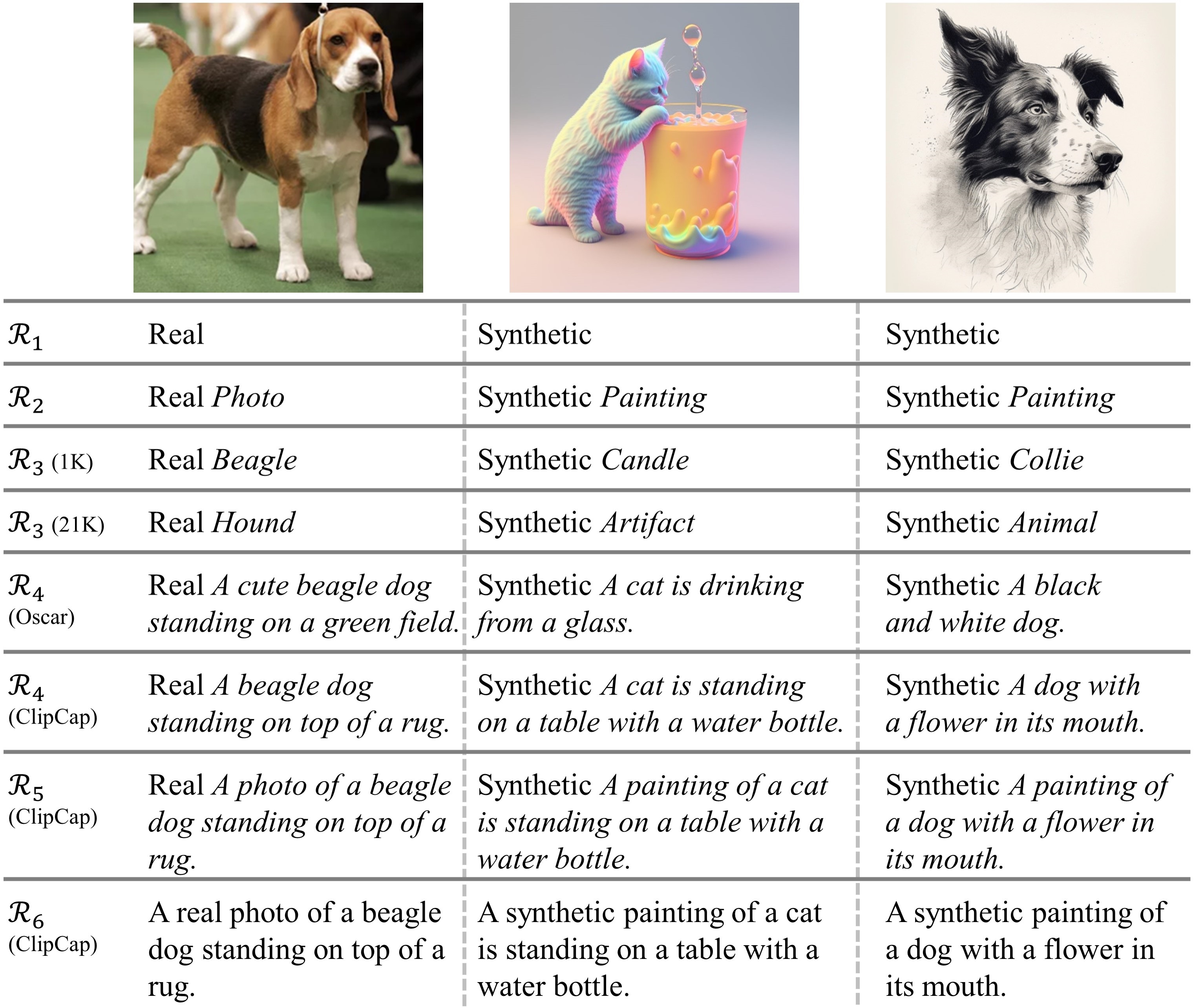}}
	\caption{Examples of different textual labels. The normal text represents the main label, while the secondary label are \textit{italic}.}
	\label{fig:text_label_demo}
\end{figure}

\paragraph{Training Paradigm}
In Table~\ref{tab:ablation}, the results of LASTED are given in row \#11. We also train the same network using a standard classification and image-image contrastive paradigm, and the results are reported in rows \#1 and \#2, respectively. More specifically, we utilize binary cross-entropy loss to supervise the image encoder, while in the image-image contrastive paradigm, we use the cyclic loss \cite{circleloss}, without involving any text encoder. It is obvious that the classification paradigm is not sufficient enough to extract generalizable representations, achieving only .842 AP. The networks trained with the image-image contrastive paradigm perform even worse, with .803 AP. It should be emphasized that the objective function of the image-image contrastive paradigm is much more difficult to be optimized, making the model easily get trapped in local optima. 

Compared to using a single-modal contrastive paradigm, the proposed LASTED establishes a language-guided contrastive based on Eqs. (1) and (2). Specifically, in rows \#3 and \#4, we ablate Eq. (1) and Eq. (2) respectively to assess the effects of constraints solely from the textual or visual perspectives. It can be seen that the rows \#3 and \#4, lacking either image or text constraints, achieve at most .848 and .857 AP respectively, significantly lower than the .943 AP attained in \#11 when both constraints are applied simultaneously. This phenomenon arises primarily because in this task, images and text do not have a one-to-one correspondence. For instance, the text ``Real'' may correspond to multiple real images. Therefore, the combined loss on the image and text axes empowers our proposed LASTED to acquire a robust multimodal representation.

In addition to the above experiments, we evaluate the zero-shot capabilities of different pre-trained paradigms. Since the objective of these pre-training paradigms is not to detect synthetic images, we use feature metrics for evaluation. Specifically, we expect that the feature similarity between two synthetic images (or real images) will be higher, otherwise lower. The results in Table~\ref{tab:zero_shot} show that CLIP \cite{clip} based on language supervision has better representation capabilities, far exceeding the MoCo \cite{moco} and MAE \cite{mae} paradigms. This also once again verifies the effectiveness of our LASTED. Note that although pre-trained CLIP already has certain zero-shot capabilities, it is still difficult to compare with supervised methods.

\begin{figure}[t!]
	\centering
	\subfloat{
		\includegraphics[width = 0.48\textwidth]{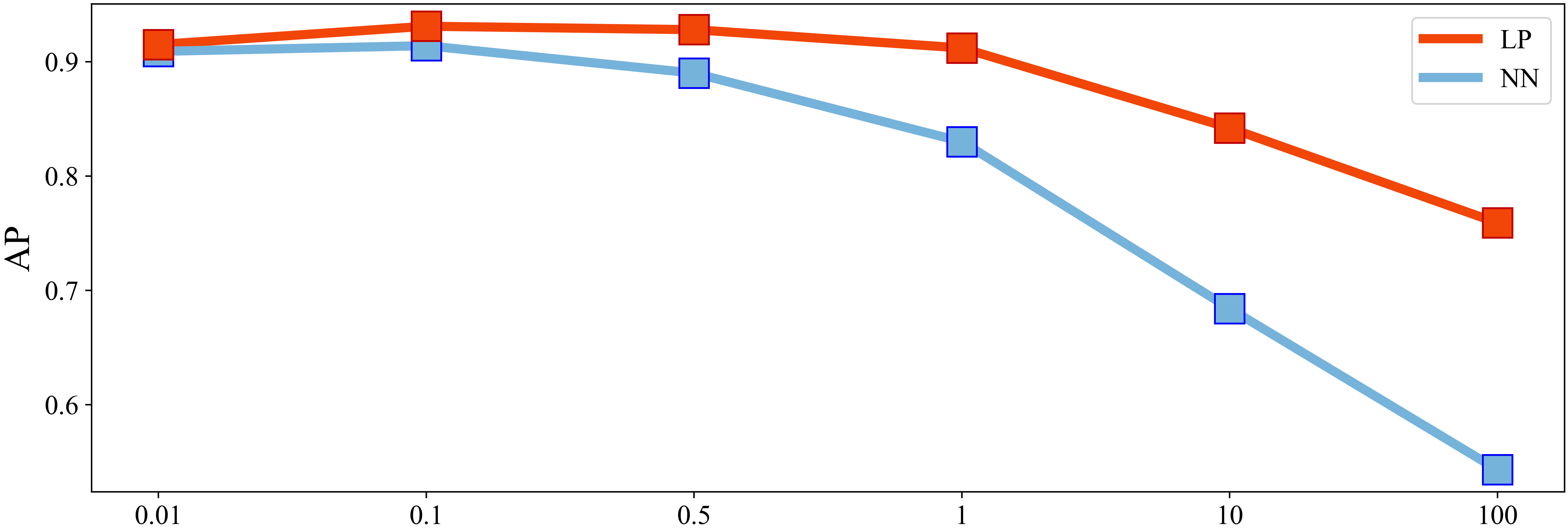}}
	\caption{The impact of the $\lambda$ over the testing set $\mathcal{T}_{wild}$.}
	\label{fig:lambda}
\end{figure}

\paragraph{Textual Label}
As mentioned in Sec.~\ref{sec:method_label}, language supervision has various textual labeling strategies that could affect the detection performance. In rows \#5$\sim$\#12, we compare eight labeling choices. Specifically, for rows \#7 and \#8, we employ ResNet \cite{resnet2016} models pre-trained on ImageNet-1K and 21K to perform object predictions as secondary labels. While the secondary labels in the \#9 and \#10 rows are derived from the semantics described by the image captioning algorithms Oscar \cite{li2020oscar} and ClipCap \cite{caption}, respectively. A demonstration of the textual labels is given in Fig.~\ref{fig:text_label_demo}. As can be seen, in the middle of the fourth row in Fig.~\ref{fig:text_label_demo}, the ``Artifact'' label cannot accurately reflect the image content. This implies that if we can acquire more precise semantic descriptions, we can achieve better improvements through the LASTED paradigm.

As can be observed, although \#5 achieves a respectable .926 AP on $\mathcal{T}_{gan}$, we can further enhance the performance by expanding textual labels (such as \#6$\sim$\#11), reaching a maximum .974 AP. Compared to classification algorithms that can only provide a single category, semantic captioning algorithms offer more information as secondary labels. Thus \#9 and \#10 can generally surpass the performance of \#7 and \#8. However, it is worth noting that the secondary labels are not entirely objective and can sometimes even result in incorrect labels. Furthermore, the results in \#12 indicate that if we naively concatenate the main and secondary labels, the image encoder would be trained from a synthetic detector to a general semantic feature extractor, where is reflected by the .659 AP achieved by NN approach. Although we can improve the performance from .659 to .856 AP through fine-tuning (LP approach), it still lags behind models trained with more fine-grained divisions, such as the \#11 with .943 AP. Overall, LP outperforms NN, primarily because LP can further refine the knowledge learned in the language encoder and provide a larger decision boundary.

\begin{figure}[t!]
	\centering
	\subfloat{
		\includegraphics[width = 0.48\textwidth]{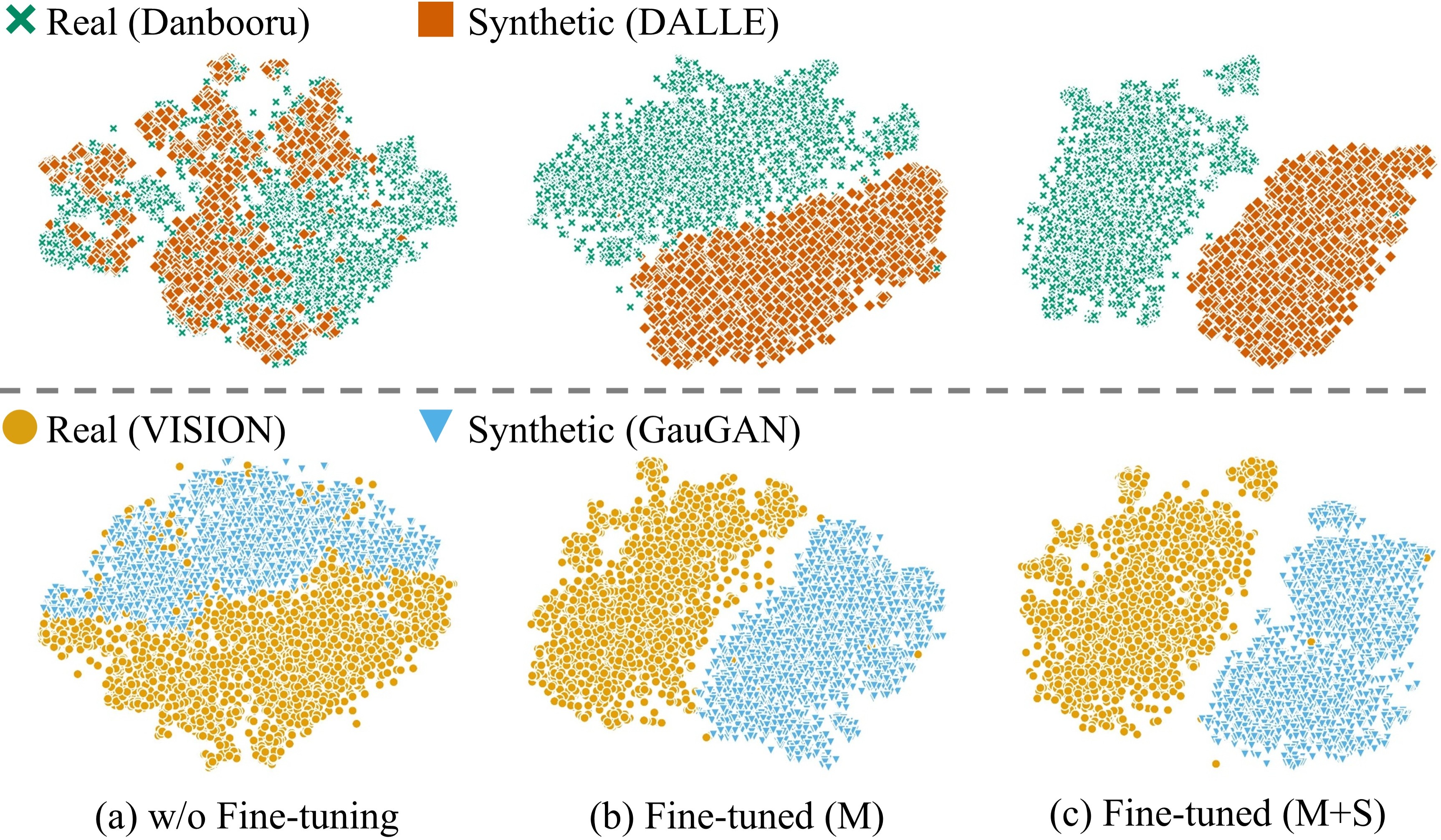}}
	\caption{Visualization of feature distributions for CLIP's ResNet50 and its variants fine-tuned under the LASTED framework with main-label (M) and secondary-label (S).}
	\label{fig:feat_vis}
\end{figure}

\paragraph{Network Architecture}
Different encoder architectures inherently possess varying representation abilities. We now evaluate the detection performance when ResNet50x64 \cite{clip}, ConvNeXt \cite{convnext}, ViT \cite{vit}, and MiT \cite{mit} are used as image encoder. Specifically, the results for ResNet50x64, ConvNeXt, ViT and MiT can be found in rows \#11, \#13, \#14, and \#15 of Table~\ref{tab:ablation}, respectively. As can be noticed, for the task of synthetic image detection, ResNet-based networks are more suitable than Transformers. As a result, we ultimately select ResNet50x64 as our image encoder. As for the text encoder, we also experiment with different architectures, such as DistilBERT \cite{DistilBERT} and ALBERT \cite{ALBERT}; however, we cannot observe noticeable performance differences. Therefore, we simply adopt Text Transformer \cite{radford2019language} as the text encoder.

\paragraph{Impact of $\lambda$}
We investigate the impact of the loss balancing parameter $\lambda$ on the final results and display the results on the $\mathcal{T}_{wild}$ dataset in Fig.~\ref{fig:lambda}. Clearly, if $\lambda$ is too large, it causes the main labels during training to lose prominence, turning the trained encoder into a general semantic extractor. Similarly, although subsequent LP can learn features that distinguish between real and synthetic, it still struggles to achieve the expected results. Therefore, we ultimately set $\lambda$ to 0.1.

\paragraph{Impact on Feature Distribution}
Fig.~\ref{fig:feat_vis} visualizes the extracted features of a CLIP pre-trained ResNet50 and its two fine-tuned variants under our LASTED: one trained with main labels and the other supplemented with secondary labels. The pre-trained model struggles under challenging cross-domain distributions (Danbooru vs. DALLE). While main-label fine-tuning yields slight gains, the model still lacks discriminative power, likely from overfitting to training domain (ImageNet vs. ProGAN). Introducing secondary labels alleviates this overfitting, enhancing cross-domain feature discrimination.

\subsection{Robustness to Post-processing}
\begin{figure}[t!]
	\centering
	\subfloat{
		\includegraphics[width = 0.49\textwidth]{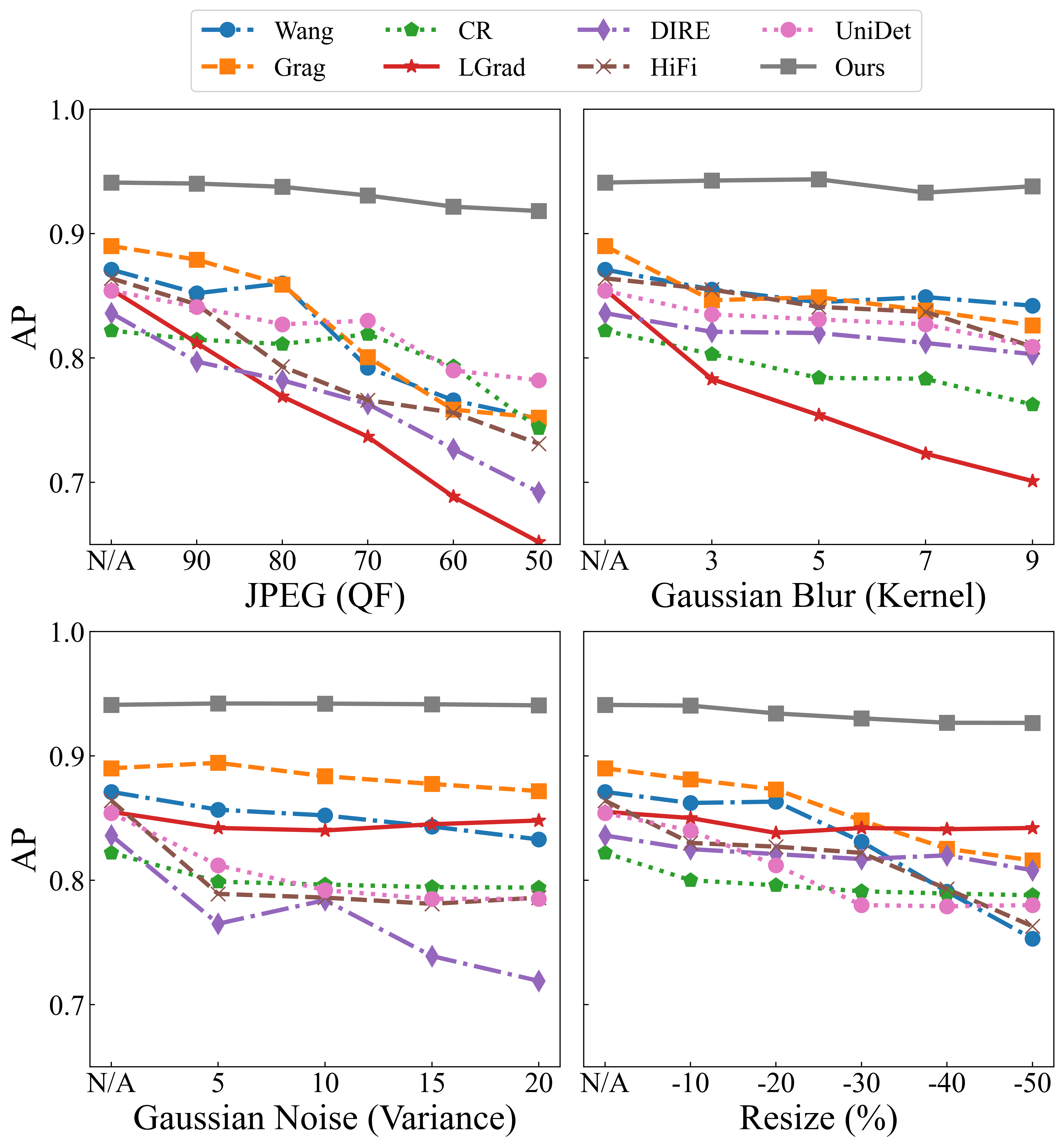}}
	\caption{Robustness evaluations against compression, blurring, noise addition, and resizing.}
	\label{fig:robustness}
\end{figure}

We also analyze the robustness of all competing detectors against post-processing operations. This is crucial because the given images under investigation may have gone through various post-processing operations. Specifically, these operations refer to actions taken after the initial image generation process, which include any user-driven edits, such as cropping, resizing, or compressing that could impact the final appearance and quality of the images. To this end, we select four commonly-used operations, including JPEG compression, Gaussian blurring, Gaussian noise, and down-sampling. We then apply them to the challenging practical dataset $\mathcal{T}_{wild}$, and show the results in Fig.~\ref{fig:robustness}. It can be seen that CR \cite{color_robust2022} and Grag \cite{grag2021} are somewhat vulnerable against JPEG compression, especially when the quality factors (QFs) are small. For instance, when QF is 50, the performance drop can be more than 10\%. Fortunately, our LASTED consistently demonstrates satisfactory robustness against these post-processing interference.           

\section{Conclusions}\label{sec:conclusion}
This paper addresses an important issue on how to improve the generalization of synthetic image detector. To this end, we propose LASTED, a language-guided contrastive learning framework with novel training paradigm, for the generalizable synthetic image detection. Extensive experiments are provided to demonstrate the strong generalizability of our proposed LASTED, which outperforms the state-of-the-art competitors by a big margin.

In our future work, we will focus on designing a robust synthetic image detector to tackle the potential challenges posed by images generated from maliciously manipulated prompts. Specifically, we aim to design multimodal adversarial training to enhance the detector's adaptability to various types of malicious attacks, whether these attacks come from image or text modalities. We hope these efforts will lead to a more robust and trustworthy synthetic image detector.

\bibliographystyle{IEEEtran}
\bibliography{ref}

\end{document}